\pdfoutput=1
\documentclass{article} % For LaTeX2e
\usepackage{iclr12submit_e,times}
\usepackage[pdftex]{graphicx}
\usepackage{amsmath}
\usepackage[square]{natbib}
\usepackage{hyperref}
\usepackage{enumitem}

\hypersetup{
    colorlinks=true,
    linkcolor=black,
    citecolor=black,
    filecolor=black,
    urlcolor=black,
}

\title{The Neural Representation Benchmark and its Evaluation on Brain and Machine}

\author{
Charles F.~Cadieu, 
Ha Hong,
Dan Yamins,
Nicolas Pinto,
Najib J. Majaj,
James J. DiCarlo
\\
McGovern Institute for Brain Research and\\
Department of Brain and Cognitive Sciences\\
Massachusetts Institute of Technology\\
Cambridge, MA 02139\\
\texttt{cadieu@mit.edu}\\
}

\nipsfinalcopy % Uncomment for camera-ready version

\begin{document}

\maketitle

\begin{abstract}
A key requirement for the development of effective learning representations is their evaluation and comparison to representations we know to be effective.  In natural sensory domains, the community has viewed the brain as a source of inspiration and as an implicit benchmark for success.  However, it has not been possible to directly test representational learning algorithms directly against the representations contained in neural systems.  Here, we propose a new benchmark for visual representations on which we have directly tested the neural representation in multiple visual cortical areas in macaque (utilizing data from~\citep{Majaj:2012ui}), and on which any computer vision algorithm that produces a feature space can be tested.  The benchmark measures the effectiveness of the neural or machine representation by computing the classification loss on the ordered eigendecomposition of a kernel matrix~\citep{Montavon:2011wp}.  In our analysis we find that the neural representation in visual area IT is superior to visual area V4, indicating an increase in representational performance in higher levels of the cortical visual hierarchy.  In our analysis of representational learning algorithms, we find that three-layer models approach the representational performance of V4 and the algorithm in~\citep{le2011building} surpasses the performance of V4.  Impressively, we find that a recent supervised algorithm~\citep{Krizhevsky:2012wl} achieves performance comparable to that of IT for an intermediate level of image variation difficulty, and surpasses IT at a higher difficulty level.  We believe this result represents a major milestone: it is the first learning algorithm we have found that exceeds our current estimate of IT representation performance.  To enable researchers to utilize this benchmark, we make available image datasets, analysis tools, and neural measurements of V4 and IT.  We hope that this benchmark will assist the community in matching the representational performance of visual cortex and will serve as an initial rallying point for further correspondence between representations derived in brains and machines.
\end{abstract}

\section{Introduction}
One of the primary goals of representational learning is to produce algorithms that learn transformations from unstructured data and produce representational spaces that are well suited to problems of interest, such as visual object recognition or auditory speech recognition.  In the pursuit of this goal, the brain and the representations that it produces has been used as a source of inspiration and even suggested as a benchmark for success in the field.  In this work, we attempt to provide a new benchmark to measure progress in representational learning with defined measures of success relative to high-level visual cortex.
% (appendix reference to various quotes in the literature)

The machine learning and signal processing communities have achieved many successes by incorporating insights from neural processing, even when a complete understanding of the neural systems was lacking.  The initial formulations of neural networks took explicit inspiration for how neurons might transform their inputs~\citep{Rosenblatt:1958jc}.  David Lowe, in his original formulation of the SIFT algorithm cites inspiration from complex cells in primary visual cortex~\citep{lowe2004distinctive} and IT cortex~\citep{lowe2000towards}.  The concepts of hierarchical processing and intermediate features also have a history of cross pollination between computer vision and neuroscience~\citep{Fukushima:1980iz,Riesenhuber1999,Stringer:2002hv,serre2007,Pinto:2009ho}.  This cross pollination has also had great influence on the field of neuroscience and has suggested ways to investigate how the brain works, suggesting specific hypotheses about its computational principles.  For the work presented here, the architectures and algorithms devised for hierarchical (deep) neural networks may serve as concrete hypotheses for the computational mechanisms used by the visual cortex to achieve fast and robust object recognition.  We believe that the neuroscience field needs more concrete hypotheses and we hope that the latest representational learning algorithms will fill that void.

How do we measure representational efficacy?  Any quantitative evaluation of progress made in representational learning must address this question.  Here we advocate for the use of ``kernel analysis,'' formulated in the works of~\citep{Braun:2006va,Braun:2008ul,Montavon:2011wp}.  We believe that kernel analysis has two main advantages.  First, it measures the accuracy of a representation as a function of the complexity of the task decision boundary.  This allows us to identify representations that achieve high accuracy for a given complexity.  This also avoids a measurement confound that arises when using cross-validated accuracy: the decision boundary's complexity and/or constraints are dependent on the size and choice of the training dataset, factors that can strongly affect accuracy scores.  By measuring how the accuracy is affected by the complexity of the decision boundary, kernel analysis allows us to explicitly take this dependency into account.  Second, kernel analysis is particularly advantageous for comparisons between models and the brain because it is robust to the number of samples used in the measurement.  While our ability to measure neural activity in the brain has increased exponentially (see ~\citep{Stevenson:2011ur} for an analysis of simultaneous recording growth rate, which is related to the number stimuli that can be measured), we are still orders of magnitude away from the dataset sizes achieved in the machine learning community.  For this reason, measures that are useful in this low-sample regime are particularly important when evaluating the performance of neural representations.  Kernel analysis exhibits this property as it converges quickly as a function of the number of samples (in our case images) used in the analysis.  Therefore, while other measures of representational efficacy may be related to kernel analysis (such as cross-validated classification accuracies, or counting the number of support vectors) we here utilize kernel analysis for its convergence properties and explicit measurement of accuracy versus complexity.
%does it make sense to reference "one-shot learning", or at any rate, the idea that human behavior seems to generalize from small numbers of stimuli in a way that many machine representations have not?  perhaps this motivation can be made more explicit...

In general, there are a number of methodologies we might consider when comparing algorithms to neural responses.  One approach is to model neural variation directly~\citep{wu2006complete}.  This approach is valid scientifically in the pursuit of understanding neural mechanisms, but it lacks a representational aspect.  For example, some details of neural activity may have no representational value, insofar as their variation does not relate to any variable we are interested in representing outside the neural mechanism.  Therefore, we seek a measure that blends the neural measurement with the representational tasks of interest.  This approach does have its downsides; most troubling of which is that we must choose a specific aspect of the world that is represented in the neural system.  We can hope that our chosen task is one that the neural system effectively represents -- ideally, one that the neural system has been optimized to represent.  A major, unaccomplished, goal of computational neuroscience is to determine the representation formed in the brain by finding the mapping between external factors and neural response.  In the methodology we propose, we do not claim to have solved the problem of choosing the aspects of the world that the brain has been optimized to represent, but we do believe we have chosen a reasonable ÔtaskÕ or aspect of the visual environment: category-level object recognition.\footnote{In relation to the scientific goal of finding those aspects of the world that the brain is representing, kernel analysis may be a way to measure which aspects of the world the brain has been optimized to represent: the attributes of the environment that the neural representation is found to perform well on, may be those aspects that the brain has been optimized to represent.  However, such an examination is beyond the scope of this paper.}

This work builds on a series of previous efforts to measure the representational efficacy of models against that of the brain.  The work of Nikolaus Kriegeskorte and colleagues, see for example~\citep{Kriegeskorte:2008vz}, examined the variation present in neural populations to visual stimuli presentations and compared this variation to the variation produced in model feature spaces to the same stimuli.  This work has influenced us in the pursuit of finding such mappings, but it has a major downside for our purposes: it does not measure the variations in the neural or model spaces that are relevant for a particular task, such as class-level object classification\footnote{However, see \citep{NikolausKriegeskorte:2008bz} and \citep{Mur:2012vq}, for discussion of methodologies to account for dissimilarity matrices by class-distance matrices.  Such a methodology will produce a single summary number, and not the accuracy-complexity curves we achieve with kernel analysis.}.  There exist a number of published accounts of neural datasets that might be useful for the type of comparison we seek~\citep{wallis1997,hung2005fast,Kiani:2007uz,Rust:2010uk,Zhang:2011ut}, but these measurements have not been released, are often made on only a handful of images, and the measures given -- typically cross-validated performance -- are not as robust to low image counts as the kernel analysis metric we use here.

In comparing algorithms to the brain, it is important to choose carefully the neural system to measure and the type of neural measurement to make.  In this work we analyze the ventral stream of macaque monkey, a non-human primate species.  Using macaque visual cortex allows us to leverage an extensive literature that includes behavioral measurements~\citep{FabreThorpe:1998te}, neural anatomy~\citep{felleman1991distributed}, extensive physiological measurements in numerous cortical visual areas, and measurements using a variety of techniques, from single cell measurements, to fMRI (for a review of high-level processing see~\citep{orban2008higher}).  These experiments indicate that macaque has visual abilities that are close to those of humans, that the ventral cortical processing stream (spanning V1, V2, V4, and IT) is relevant for object recognition, and that multi-unit recordings in high-level visual areas exhibit responses that are increasingly robust to object identity preserving variations (for a review see~\citep{DiCarlo:2012em}).%While we believe that carrying out our protocol utilizing fMRI in humans or macaque would be instructive, we believe that at this point in time, cortical representation is better elucidated at the level of multi-unit recordings in macaque.

With these considerations in mind, we describe a neural representation benchmark that may be used to judge the representational efficacy of representational learning algorithms.  Importantly, we present a measurement of visual areas V4 and IT in macaque cortex on this benchmark.  These measurements allow researchers to test their algorithms against a known, high-performing representation.  They may also provide an evaluation and thus facilitate a long sought goal of artificial intelligence: to achieve representations as effective as those found in the brain.  Our preliminary evaluation of machine representations indicates that we may be coming close to this goal.

The paper is organized as follows. In the Methods section we describe the images and task we use, the neural measurements, the use of kernel analysis, and our suggested protocol for measuring algorithms.  In the Results section we provide the kernel analysis measurement on V4 and IT, on a number of control models, and on some recently published high-performing neural network models.  We conclude with a discussion of additional aspects of the neural system that will need to be investigated to ultimately conclude that representational learning algorithms are as effective as the brain.

\section{Methods}
The proposed benchmark utilizes an image dataset composed of seven object classes and is broken down into three levels of variation, which present increasing levels of difficulty.  We measure the representational efficacy of a feature space using kernel analysis, which measures the classification loss under an eigendecomposition of the representation's kernel matrix (kernel PCA).%In this section we describe the image generation process, the computation of kerneled analysis for the suggested testing protocol, and describe the neural data collection process and the models we evaluate.

\subsection{Image dataset generation} \label{sec:imageset}
\begin{figure}[h]
\begin{center}
\includegraphics[width=\linewidth]{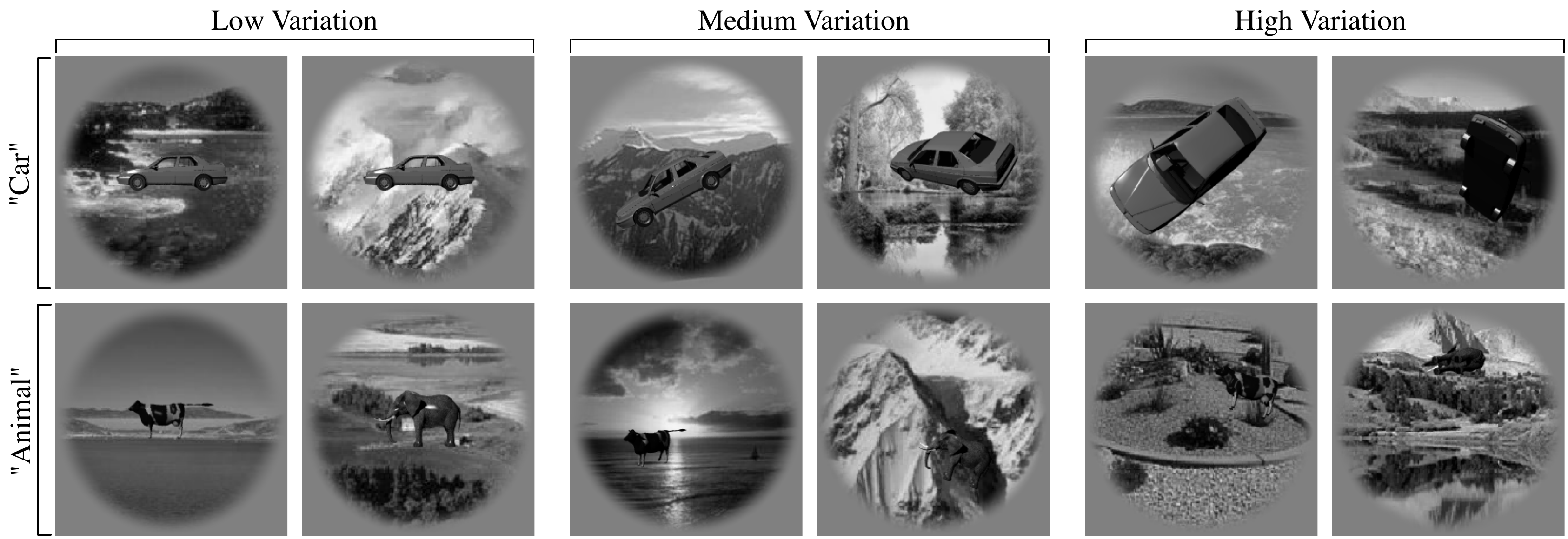}
\end{center}
\caption{Example testing images for each variation level.  For each variation level, Low, Medium and High Variation, we show two example images from the Car class and two example images from the Animal class.  The car images shown all contain the same object instance, thus showing the image variability due only to the variation parameters and backgrounds.  The animal images contain either a cow object instance or an elephant object instance, thus showing variability due to exemplar, variation parameters, and background.}
\label{fig:images}
\end{figure}

For the representational task, we have chosen class-level object recognition under the effect of image variations due to object exemplar, geometric transformations (position, scale, and rotation/pose), and background.  The task is defined through an image generation process.  An image is constructed by first choosing one of seven categories, then one of seven object exemplars from that category, then a randomly chosen background image, and finally the variation parameters drawn from one of three distributions.  The three different variation parameter distributions systematically increase the degree of variation that is sampled, from Low Variation, which presents objects at a fixed position, scale, and pose, to Medium Variation, to High Variation, which presents objects at positions spanning the image, under multi-octave scale dilation, and from a wide range of poses.  Example images for each variation level are shown in Figure~\ref{fig:images}.

The resulting image set has several advantages and disadvantages.  Advantageously, this procedure eliminates dependencies between objects and backgrounds that may be found in real-world images~\citep{Oliva:2007ui}, and introduces a controlled amount of variability or difficulty in the task, which has been used to produce image datasets that are known to be difficult for current algorithms~\citep{Pinto:2008gj,Pinto:2010hvm,Pinto:2011iw}.  While the resulting images may have an artificial quality to them, having such control allows us to scientifically investigate neural coding in relation to these parameters.  The disadvantages of using this image set are that it does not expose contextual effects that are present in the real world and may be used by both neural and machine systems, and we do not (currently) include other relevant variations, e.g. lighting, texture, natural deformations, or occlusion.  We view these disadvantages as opportunities for future datasets and neural measurements.
%Define the classification problem and stimulus generation (describe image generation process, mainly reference another paper?)
%	advocate for certain advantages:
%		no correlation between objects and backgrounds
%		controlled variability along certain dimensions: category exemplar, pos/scale/rot
%	bring up downsides:
%		does not expose contextual effects present in the real world (Torralba)
%		No deformable objects
%		Lacks other sources of variation: texture, occlusion

\subsection{Kernel analysis methodology}
In measuring the efficacy of a representation we seek a measure that will favor representations that allow for a simple task solution to be learned.  For this measure, we turn to the work presented in~\citep{Montavon:2011wp}, which is based on theory presented in~\citep{Braun:2006va}, and~\citep{Braun:2008ul}.  We provide a brief description of this measure and refer the reader to those references for additional details and justification.

The measurement procedure, which we refer to here as \emph{kernel analysis}, utilizes kernel principal component analysis to determine how much of the task in question can be solved by the leading kernel principal components.  Kernel principal components analysis will decompose the variation in the representational space due to the stimuli in question.  A \emph{good} representation will have high variability in relation to the task in question.  Therefore, if the leading kernel principal components are effective at modeling the task, the representational space is effective for that task.  In contrast, an ineffective representational space will have very little variation relevant for the task in question and variation relevant for the task is only contained in the eigenvectors corresponding to the smallest eigenvalues of the kernel principal component analysis.  Intuitively, a good representation is one that learns a simple boundary from a small number of randomly-chosen examples, while a poor representation makes a more complicated boundary, requiring many examples to do so.

Following~\citep{Montavon:2011wp}, kernel analysis consists of estimating the $d$ first components of the kernel feature space and fitting a linear model on this low-rank representation to minimize the loss function for the task.  The subspaces formed by the $d$ first components controls the complexity of the model and the accuracy is measured by the loss in that subspace $e(d)$.  We refer to the dimensionality of the subspace $d$ as the \emph{complexity} and $1 - e(d)$ as the \emph{accuracy}.  Thus, the curve $1 - e(d)$ provides us with a measurement of the accuracy as a function of the model complexity for the given representational space.  The curves produced by different representational spaces will inform us about the simplicity of the task in that representational space, with higher curves indicating that the problem is simpler for the representation.

One of the advantages of kernel analysis is that the kernel PCA method converges favorably from a limited number of samples.  \citet{Braun:2008ul} show that the kernel PCA projections obtained with a finite and typically small number of samples $n$ (images) are close with multiplicative errors to those that would be obtained in the asymptotic case where $n \mapsto \infty$.  This result is especially important in our setting as the number of images we can reasonably obtain from the neural measurements is comparatively low.  Therefore, kernel analysis provides us with a methodology for assessing representational effectiveness that has favorable properties in the low image sample regime, here thousands of images.

We next present the specific computational procedure for computing kernel analysis.  Given the learning problem $p(x,y)$ and a set of $n$ data points $\{(x_1,y_1),..., (x_n,y_n)\}$ drawn independently from $p(x,y)$ we evaluate a representation defined as a mapping $x\mapsto \phi(x)$.  For our case, the inputs $x$ are images, the $y$ are category labels, and the $\phi$ denotes a feature extraction process.

As suggested by~\citep{Montavon:2011wp}, we utilize the Gaussian kernel because this kernel implies a smoothness of the task of interest in the input space~\citep{Smola:1998dq}.  We compute the kernel matrix $K_\sigma$ associated to the data set as
\begin{equation}
K_\sigma =  \begin{pmatrix}
k_\sigma(\phi(x_1),\phi(x_1)) & ... & k_\sigma(\phi(x_1),\phi(x_n)) \\
 \vdots &  &\vdots \\
k_\sigma(\phi(x_n),\phi(x_1)) & ... & k_\sigma(\phi(x_n),\phi(x_n)) \\
\end{pmatrix},
\end{equation}
where the standard Gaussian kernel is defined as $k_\sigma (x, x') = \exp ( - || x - x' ||^2 / 2\sigma^2)$.

We perform an eigendecomposition of $K_\sigma$ where the eigenvectors $u_1, ..., u_n$ are sorted in decreasing magnitude of their corresponding eigenvalues $\lambda_1, ..., \lambda_n$:
\begin{equation}
K_\sigma = (u_1 | ... | u_n) \cdotp \mathrm{diag} (\lambda_1, ..., \lambda_n) \cdotp (u_1 | ... | u_n)^T .
\end{equation}

Let $\hat U_d = (u_1 | ... | u_d)$ and $\hat \Lambda_d = \mathrm{diag} ( \lambda_1, ..., \lambda_d)$ be the $d$-dimensional approximation of the eigendecomposition.  Note that we have dropped, for the moment, the dependency on $\sigma$.  We then solve the learning problem using a linear model in the corresponding subspace.  For our problem we find the least squares solution to the multi-way regression problem denoted as $\Theta^*_d$ and defined as
\begin{equation}\label{eq:approxdiff}
\Theta^*_d = \operatorname{argmin}_{\Theta} ||\hat U_d \Theta - Y ||^2_F = \hat U_d^T Y.
\end{equation}

The resulting model prediction is then $\hat Y_d = \hat U_d \Theta^*_d$.  The resulting loss, with dependence on $\sigma$ is
\begin{equation}\label{eq:loss}
e(d,\sigma) = \frac{1}{n} || \hat Y_d - Y ||^2_F.
\end{equation}
To remove the dependence of the kernel on $\sigma$ we find the value that minimizes the loss at that dimensionality $d$: $e(d) = \operatorname{argmin}_{\sigma} e(d,\sigma)$.  Finally, for convenience we plot \emph{accuracy} ($1 - e(d)$) against normalized \emph{complexity} ($d / D$), where $D$ is total dimensionality.

Note that we have chosen to use a squared error loss function for our multi-way classification problem.  While it might be more appropriate to evaluate a multi-way logistic loss function, we have chosen to use the least-squares loss for its computational simplicity, because it provides a stronger requirement on the representational space to reduce variance within class and to increase variance between classes, and it allows us to distinguish representations that may be identical in terms of separability for a certain dimensionality $d$ but still have differences in their feature mappings.  The kernel analysis of deep Boltzmann machines in~\citep{Montavon:2012ub} also uses a mean squared loss function in the classification problem setting.
%	http://jmlr.csail.mit.edu/papers/volume7/braun06a/braun06a.pdf
%	http://jmlr.csail.mit.edu/papers/volume9/braun08a/braun08a.pdf
%	http://jmlr.csail.mit.edu/papers/volume12/montavon11a/montavon11a.pdf

In the discussion above, $Y = (y_1, \ldots, y_n)$ represents the vector of task labels for the images $(x_1, \ldots, x_n)$.  In our specific case, the $y_i$ are category identity values, and are assumed to be discrete binary values in equations \ref{eq:approxdiff} and \ref{eq:loss} above.  To generalize to the case of multiway categorization, we use a version of the common one-versus-all strategy.  Assuming $k$ distinct categories, we form the label matrix
\begin{equation}
\mathbf{Y} = (y_{ij}) = \begin{cases} 
                    1 \mbox{ if image $x_i$ is in category $j$} \\
                    0 \mbox{ otherwise} 
                \end{cases}
\end{equation}
where $j \in [1, \ldots, k]$.  Then for each category $j$, we compute the per-class prediction $Y_d^j$ by replacing $Y$ in equations \ref{eq:approxdiff} and \ref{eq:loss} with $Y_j$, the $j$-th column of $\mathbf{Y}$.  The overall error is then the average over classes of the per-class error, e.g.
\begin{equation}
e(d,\sigma) = \langle e_j(d, \sigma) \rangle_j = \left \langle \frac{1}{n} || \hat Y_d^j - Y_j||^2_F \right \rangle_j.
\end{equation}
Minimization over $\sigma$ then proceeds as in the binary case. 

%%%or, if preferred, this somewhat more technically "coherent" but abstract version of the multi-way explanation: 
%In the discussion above, the labels $Y$ were assumed to be scalar values, but it is easy to generalize to vector-valued labels by assuming a label matrix $\mathbf{Y}$ whose rows correspond to datapoints and columns correspond to label components.  We then get vector-valued versions of the regression and prediction whose $j$-th components are computed by replacing $Y$ in equations \ref{eq:approxdiff} and \ref{eq:loss} with $Y_j$, the $j$-th column of $\mathbf{Y}$.  The overall loss now simply reduces to the mean over component losses, e.g. 
%\begin{equation}
%e(d,\sigma) = \langle e_j(d, \sigma) \rangle_j = \left \langle \sum_j \frac{1}{n} || \hat Y_d^j - Y_j||^2_F \right \rangle_j.
%\end{equation} 
%Minimization over $\sigma$ then procedes as in the scalar case. 
%In our specific case of multi-way categorization, we use a version of the common one-versus-all strategy, e.g. assuming $k$ distinct object categories, we form the label matrix
%\begin{equation}
%\mathbf{Y} = (y_{ij}) = \begin{cases} 
%                    1 \mbox{ if image $x_i$ is in category $j$} \\
%                    0 \mbox{ otherwise} 
%                \end{cases}
%\end{equation}
%where $j \in [1, \ldots, k]$.
\subsection{Suggested protocol}
To evaluate both neural representations and machine representations we measure the kernel analysis curves and area under the curves (KA-AUC) for each variation.  The testing image dataset consists of seven object classes with seven instances per object class, broken down into three levels of variation, with 490 images in Low Variation, 1960 in Medium Variation, and 1960 in High Variation.  The classes are Animals, Cars, Chairs, Faces, Fruits, Planes and Tables.  To measure statistical variation due to subsampling of image variation parameters we evaluate 10 pre-defined subsets of images, each taking 80\% of the data from each variation level.  Within each subset we equalize the number of images from each class.  For each representation, we maximize over the values of the Gaussian kernel $\sigma$ parameter chosen at 10\%, 50\%, and 90\% quantiles in the distance distribution.  For each variation level and representation, this procedure produces a kernel analysis curve and AUC for each of the subsets, and we compute the mean and standard deviation of the AUC values.

We also provide a training dataset that may be used for model selection.  This dataset allows both unsupervised and supervised training of representational learning algorithms.

\subsection{Provided data and tools}
To allow researchers to utilize this dataset we provide the following tools and downloadable data:
\begin{itemize}[topsep=0pt,leftmargin=.5cm]
\item Testing images: a set of images containing seven object classes with seven instances per object class, broken down into three levels of variation, with 490 images in Low Variation, 1960 in Medium Variation, and 1960 in High Variation.  The classes are Animals, Cars, Chairs, Faces, Fruits, Planes and Tables.  Computing features on this set of images is sufficient to evaluate an algorithm.  Each image is grayscale and 256 by 256 pixels.  To prevent over-fitting, candidate algorithms should \emph{not} be trained on this dataset, and any parameter estimation involved in model selection should be estimated independently of these testing images.
\item Training images: a set of 128,000 images consisting of 16 object classes with 16 object instances per object class, these images are produced from a similar rendering procedure as the testing image set.  The training set contains no specific constituent objects or background images in common with the testing set, but it does have new objects in each of the original seven categories, in addition to 9 new categories.  This image set \emph{can} therefore be used for independent model selection and learning, using either supervised for unsupervised methods, see Appendix A.  Use of this training set is, however, optional.
\item Testing set kernel analysis curves and KA-AUC values for V4 and IT.
\item Tools to evaluate kernel analysis from features produced by a model to be tested.
\end{itemize}
These tools and datasets can be found at: \small{\url{http://dicarlolab.mit.edu/neuralbenchmark}}\normalsize.

\subsection{Neural data collection}

We collected 168 multi-unit sites from IT cortex and 128 multi-unit sites from V4.  To form the neural feature vectors for IT and V4 we normalized responses by background firing rate and by variance within a presentation of all images within a variation.  See Appendix B for details.  This post-processing procedure has been shown to account for human performance~\citep{Majaj:2012ui} (also see \citep{hung2005fast,Rust:2010uk} for results utilizing a similar procedure).

\subsection{Machine representations}
We evaluate a number of machine representations from the literature, including several recent best of breed representational learning algorithms and visual representation models, as well as a feed-forward three layer hierarchical model optimized on the training set. 

%\subsubsection{Pixels}
%
\textbf{V1-like} We evaluate the V1-like representation from Pinto et al.'s V1S+~\citep{Pinto:2008gj}.  This model attempts to capture a first-order account of primary visual cortex (V1).  It computes a collection of locally-normalized, thresholded Gabor wavelet functions spanning orientation and frequency.  This model is a simple, baseline biologically-plausible representation, against which more sophisticated representations can be compared.
%L2 models ?
%
%\textbf{HT-L3-1st FG 2011} We evaluate the highest performing HT-L3 model produced by the procedure described in~\citep{Cox:2011wj}.  The model is a three layer model in which each layer sequentially performs local filtering, thresholding and/or saturation, pooling and normalization.  The model parameters are sampled and the top performing model on the Labeled Faces in the Wild View 1 dataset, a face verification task, is selected.  We evaluate the top level output features here.  Note that the face verification task is a very different task than the one we evaluate in the classification protocol.

\textbf{High-throughput L3 model class (HT-L3)} We evaluate the same three layer hierarchical convolutional neural net model class described in \citep{Pinto:2009ho} and \citep{Cox:2011wj}, the ``L3 model class".  Each model in this class, is a three layer model in which each layer sequentially performs local filtering, thresholding, saturation, pooling, and normalization.  To choose a high performing model from this class, we performed a high-throughput search of the parameter space, using kernel-analysis performance on the provided training image set as the optimization criterion.  The top performing model on the training set is then evaluated on the testing set (Top HT-L3).  See Appendix A for further details.

\textbf{Coates et al. NIPS 2012} We evaluate the unsupervised feature learning model in~\citep{Coates:2012wm}, which learns 150,000 features from millions of unlabeled images collected from the Internet.  We evaluate the second layer ``complex cells,'' a 10,000 dimensional feature space, by rescaling the input images to 96 by 96 pixels and computing the model's output on a 3 by 3 grid of non-overlapping 32 by 32 pixel windows.  The resulting output is 90,000 dimensional.

\textbf{Le et al. ICML 2012} We evaluate the model in~\citep{le2011building}, which is a hierarchical locally connected sparse auto encoder with pooling and local contrast normalization and is trained unsupervised from a dataset of 10 million images downloaded from the Internet and fine-tuned with ImageNet images and labels.  We use the penultimate layer outputs (69696 features) of the network for the feature representation (the layer before class-label prediction).  Images are resized to the model's input dimensions, here 200 by 200 pixels.

\textbf{Krizhevsky et al. NIPS 2012 (SuperVision)} We evaluate the deep convolutional neural network model `SuperVision' described in~\citep{Krizhevsky:2012wl}, which is trained by supervised learning on the ImageNet 2011 Fall release ($\sim$15M images, 22K classes) with additional training on the LSVRC-2012 dataset (1000 classes).  The authors computed the features of the penultimate layer of their model (4096 features) on the testing images by cropping out the center 224 by 224 pixels (this is the input size to their model).  This mimics the procedure described in~\citep{Krizhevsky:2012wl}, in which this feature is fed into logistic regression to predict class labels.
%1. I actually cropped out the center 224x224 patch of the images you
%gave me (instead of resizing them to 224x224). I know this contradicts
%what I told you earlier, but I found this to be slightly easier to do
%and forgot to tell you about it.
%
%2. The features I sent you were computed from a model trained on
%ImageNet 2011 Fall release (~15M images) and then trained further on
%ILSVRC-2012 (not ILSVRC-2010).

\section{Results}
\subsection*{Evaluation of neural representations}
\begin{figure}[t]
\begin{center}
\includegraphics[width=.58\linewidth]{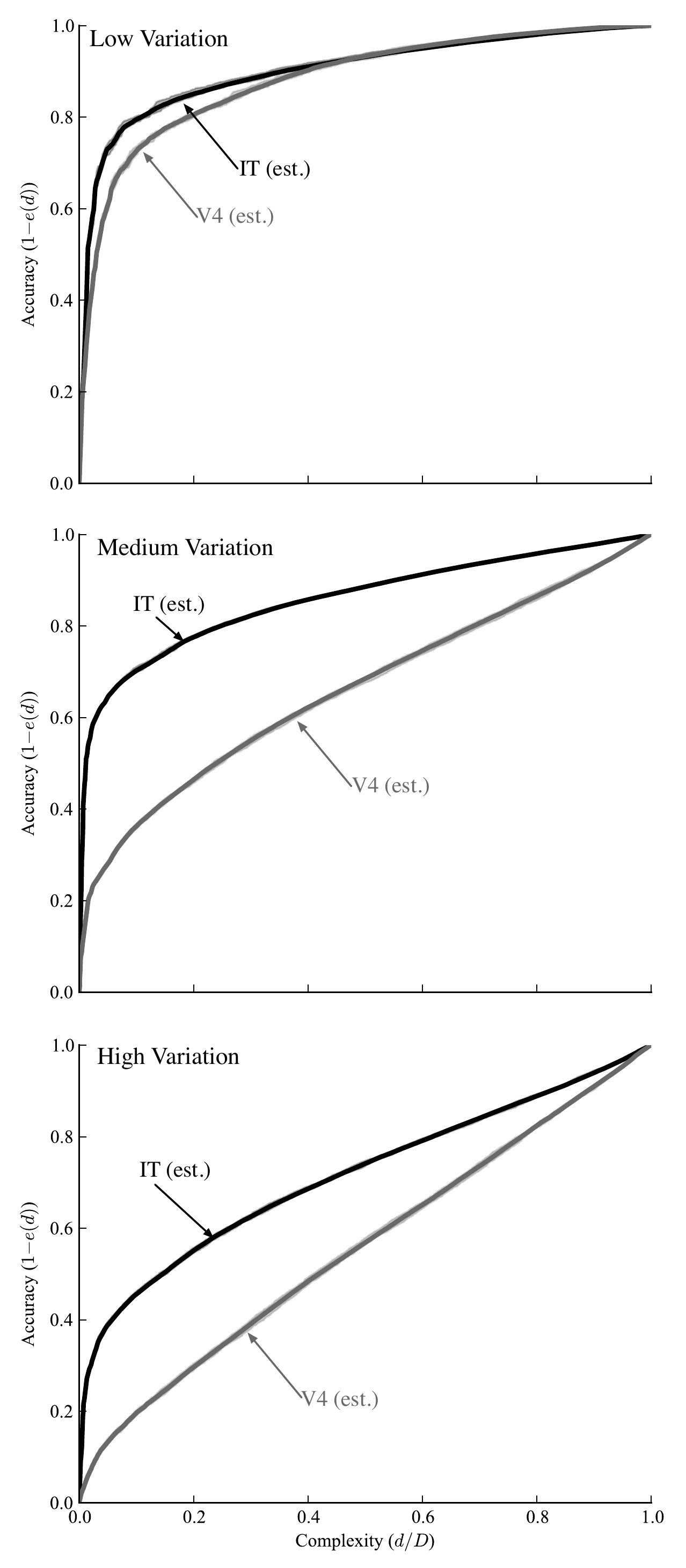}
\end{center}
\caption{Kernel analysis curves of V4 and IT.  Each panel shows the kernel analysis curves for each variation level.  Accuracy, one minus loss ($1 - e(d)$), is plotted against complexity, the normalized dimensionality of the eigendecomposition ($d/D$).  Shaded regions indicate the maximum and minimum accuracy obtained over testing subsets, which are often smaller than the line thickness.}
\label{fig:neural_curves}
\end{figure}
%\begin{figure}[h]
%\begin{center}
%%\includegraphics[width=\linewidth]{figures/ICLR_Figure_4.pdf}
%\end{center}
%\caption{(Suggest move to appendix) breakdown of all 1-vs-all curves for each variation (7x3=21 curves for V4 and IT)}
%\label{fig:neural_curves}
%\end{figure}
In Figure~\ref{fig:neural_curves} we present kernel analysis curves obtained from the measured V4 and IT neural populations for each variation level.  KA-AUC values for V4 are 0.88, 0.66, and 0.56, and for IT are 0.90, 0.86, and 0.72, for Low, Medium, and High Variation, respectively.  For each variation level, our bootstrap analysis indicates that the KA-AUC measurements between IT and V4 are significantly different (see Table~\ref{table:ka_results}).

At Low Variation there is not a large difference between V4 and IT.  This might be expected, as this variation level does not test for variability due to scale, position, or pose, which are variations that the neural responses in IT are more tolerant to than in V4.  The higher variation sets, Medium and High Variation, show increased separation between V4 and IT, and reduced performance for both representations, indicating the increased difficulty of the task under these representations.  However, the IT representation maintains high accuracy at low complexity even in the High Variation condition.  The IT representation under Medium and High Variation shows a sharp increase in accuracy at low complexity, indicating that the IT representation is able to accurately capture the class-level object recognition task with a simple decision boundary.  Note that these kernel analysis measurements are only our current estimate of representation in V4 and IT.  We discuss the limitations of these estimates in the discussion section and provide an extrapolation in Appendix C.

\subsection*{Evaluation of machine representations}
\begin{figure}[ht]
\begin{center}
\includegraphics[width=.58\linewidth]{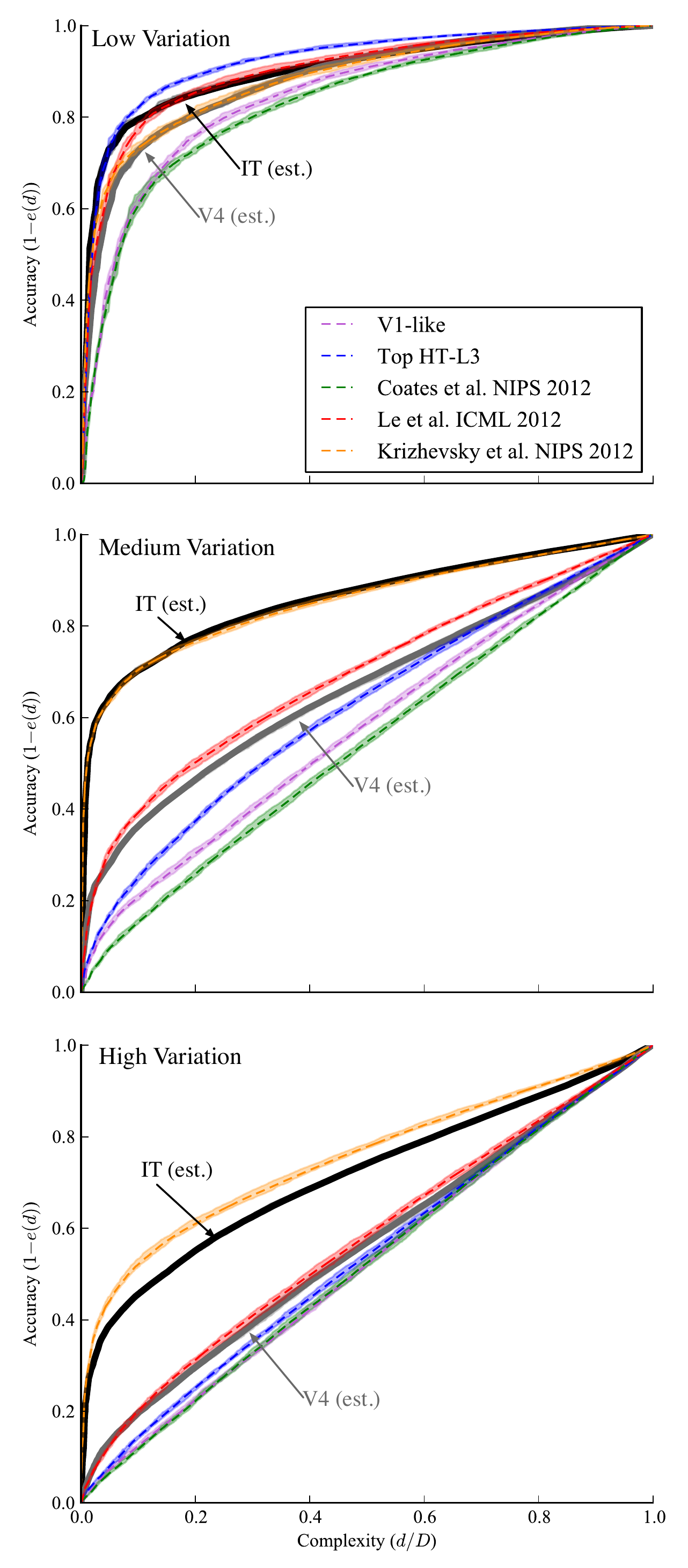}
\end{center}
\caption{Kernel analysis curves of brain and machine.  ``V4 (est.)'' and ``IT (est.)'' are brain representations and all others are machine representations.  Each panel shows the kernel analysis curves for each variation level.  Accuracy, one minus loss ($1 - e(d)$), is plotted against complexity, the normalized dimensionality of the eigendecomposition ($d/D$).  Shaded regions indicate the maximum and minimum accuracy obtained over testing subsets, which are often smaller than the line thickness.}
\label{fig:ka_machines}
\end{figure}

In Figure~\ref{fig:ka_machines} we present the kernel analysis evaluation for the machine representations we have evaluated along with the neural representations for comparison.  The corresponding KA-AUC numbers are presented in Table~\ref{table:ka_results}.  The V1-like model shows high accuracy at low complexity on Low Variation but performs quite poorly on Medium Variation and High Variation, indicating that these tasks are interesting tests of the object recognition problem.  The Top HT-L3 model is the highest performing representation at Low Variation and achieves performance that approaches V4 on Medium Variation and High Variation.  The model presented in~\citep{Coates:2012wm} performs similarity to the V1-like model on all variation levels.  This low performance may be due to the large variety of images this model was trained on, its relatively shallow architecture, and/or the mismatch in our testing image size and the 32 by 32 pixel patches of the base model.  The model presented in~\citep{le2011building} performs comparably to IT on Low Variation, and surpasses V4 at Medium and Variations.

The model in~\citep{Krizhevsky:2012wl} performs comparably to V4 at Low Variation, nearly matches the performance of IT at Medium Variation, and surpasses IT representation on High Variation.  Interestingly, this model matches IT performance at Medium Variation across the entire complexity range and exceeds it across the complexity range at High Variation.  We view this result as highly significant as it is the first model we have measured that matches our current estimate of IT representation performance at Medium Variation and surpasses it at High Variation.

%Table of results (Table 1)
%==============================
%=====  Var00
%IT Cortex  k=0.9008 (2.4e-03)
%V4 Cortex  k=0.8814 (2.0e-03)
%V1-like  k=0.8427 (2.0e-03)
%Top HT-L3  k=0.9221 (1.4e-03)
%Coates et al. NIPS 2012  k=0.8298 (1.5e-03)
%Le et al. ICML 2012  k=0.8990 (2.4e-03)
%Krizhevsky et al. NIPS 2012  k=0.8829 (2.6e-03)
%==============================
%=====  Var03
%IT Cortex  k=0.8584 (1.2e-03)
%V4 Cortex  k=0.6614 (3.2e-03)
%V1-like  k=0.5739 (2.9e-03)
%Top HT-L3  k=0.6203 (1.8e-03)
%Coates et al. NIPS 2012  k=0.5391 (3.0e-03)
%Le et al. ICML 2012  k=0.6904 (2.5e-03)
%Krizhevsky et al. NIPS 2012  k=0.8538 (2.0e-03)
%==============================
%=====  Var06
%IT Cortex  k=0.7152 (2.2e-03)
%V4 Cortex  k=0.5580 (3.2e-03)
%V1-like  k=0.5189 (2.0e-03)
%Top HT-L3  k=0.5328 (1.7e-03)
%Coates et al. NIPS 2012  k=0.5192 (2.9e-03)
%Le et al. ICML 2012  k=0.5700 (3.0e-03)
%Krizhevsky et al. NIPS 2012  k=0.7521 (3.0e-03)
\begin{table}
\caption{Kernel analysis results. For each representation we measure the KA-AUC at each variation level for each testing subset.  The means over testing subsets are given in the table, with standard deviations in parentheses.  Top performing models are highlighted.  Note that our measurements of IT and V4 Cortex are our current best estimates (est.) and are subject to experimental limitations.}
\begin{center}
ÊÊÊÊ\begin{tabular}{ | l || c | c | c |}
\hline
ÊÊÊÊÊÊÊÊÊÊÊÊÊÊÊÊÊÊÊÊÊÊÊÊÊÊ & Low VariationÊÊÊÊÊÊÊÊÊÊÊÊÊÊÊ & Medium VariationÊÊÊÊÊÊÊ & High VariationÊÊÊÊÊÊÊÊÊÊÊÊÊÊÊ \\ \hline
ÊÊÊÊÊÊÊÊIT~Cortex~(est.)ÊÊÊÊÊÊÊÊÊÊÊÊÊ & 0.90 \footnotesize{(2.4e-03)}         & \textbf{0.86} \footnotesize{(1.2e-03)} & 0.72 \footnotesize{(2.2e-03)} \\ \hline
ÊÊÊÊÊÊÊÊV4~Cortex~(est.)ÊÊÊÊÊÊÊÊÊÊÊ & 0.88 \footnotesize{(2.0e-03)}        & 0.66 \footnotesize{(3.2e-03)}              & 0.56 \footnotesize{(3.2e-03)} \\ \hline
ÊÊÊÊÊÊÊÊV1-likeÊ            ÊÊÊÊÊÊÊÊÊÊÊÊ& 0.84 \footnotesize{(2.0e-03)}        & 0.57 \footnotesize{(2.9e-03)}               & 0.52 \footnotesize{(2.0e-03)} \\ \hline
ÊÊÊÊÊÊÊÊTop HT-L3ÊÊÊÊÊÊÊÊÊÊÊÊ & \textbf{0.92} \footnotesize{(1.4e-03)} & 0.62 \footnotesize{(1.8e-03)}              & 0.53 \footnotesize{(1.7e-03)} \\ \hline
%ÊÊÊÊÊÊÊÊHT-L3-1st FG 2011ÊÊÊÊÊÊÊÊÊÊÊÊ & \textbf{0.94 (8.5e-04)} & 0.69 (2.9e-03) & 0.56 (3.2e-03) \\ \hline
ÊÊÊÊÊÊÊÊCoates et al. NIPS 2012ÊÊÊÊ & 0.83 \footnotesize{(1.5e-03)} & 0.54 \footnotesize{(3.0e-03)}            & 0.52 \footnotesize{(2.9e-03)} \\ \hline
ÊÊÊÊÊÊÊÊLe et al. ICML 2012ÊÊÊÊÊÊÊÊ & 0.90 \footnotesize{(2.4e-03)}      & 0.69 \footnotesize{(2.5e-03)}           & 0.57 \footnotesize{(3.0e-03)} \\ \hline
ÊÊÊÊÊÊÊÊKrizhevsky et al. NIPS 2012 & 0.88 \footnotesize{(2.6e-03)} & 0.85 \footnotesize{(2.0e-03)}         & \textbf{0.75} \footnotesize{(3.0e-03)} \\ \hline
\end{tabular}
\end{center}
\label{table:ka_results}
\end{table}

\section{Discussion}
There are a number of issues related to our measurement of macaque visual cortex, including viewing time, behavioral paradigm, neural subsampling, and mapping the neural recording to a neural feature, that will be necessary to address in determining the ultimate representational measurement of macaque visual cortex.  The presentation time of the images shown to the animals was intentionally brief (100 ms), but is close to typical fixation time ($\sim$200 ms).  Therefore, it will be interesting to measure how the neural representational space changes with increased viewing time, especially considering that natural viewing conditions typically allow for longer fixation times and multiple fixations.  Another aspect to consider is that animals are engaged in passive viewing during the experimental procedure.  Does actively performing a task influence the neural representation?  This question may be related to what are commonly referred to as attentional phenomena [e.g. biased competition].  Current experimental techniques only allow us to measure a small portion of the neurons in a cortical area.  While our analysis in Appendix C suggests that we are reaching saturation in our estimate of KA-AUC with our current neural sample, our sample is biased spatially on the cortical sheet because of our use of electrode grids.  This bias likely leads to an underestimate of KA-AUC.  Finally, the Ôneural codeÕ is a topic of heated debate in the neuroscience community and the mapping from multi-unit recordings to the neural feature vector we have used for our analysis is only one possible mapping.  Importantly, this mapping has been shown to account for human behavioral performance~\citep{Majaj:2012ui}.  However, as we gain more knowledge about cortical processing (such as intrinsic dynamics~\citep{Canolty:2010ts}) our best guess at the neural code may evolve and update our neural representation benchmark accordingly.

Another aspect that our measurement does not address is the direct impact of visual experience on the representations observed in IT cortex.  Interestingly, the macaques involved in these studies have had little or no real-world experience with a number of the object categories used in our evaluation, though they do benefit from millions of years of evolution and years of postnatal experience.  However, learning effects in adult IT cortex are well observed~\citep{Kobatake:1998vc,Baker:2002bt,Sigala:2002ta}, even with passive viewing~\citep{Li:2010it}.  Remaining unanswered questions are: how has the exposure during the experimental protocol affected the neural representation, and could the neural representation be further enhanced with increased exposure?

A related question: is the training set we have provided sufficient to achieve IT level performance on the testing set?  We do not have a positive example of such transfer, and we expect that algorithms leveraging massive amounts of visual data may produce the best results on the testing set.  Such algorithms, and their data dependence, will be informative.  Furthermore, to what extent do we need to build additional structure into our representations and representational learning algorithms to achieve representations equivalent to those found in the brain?

Could human neural representation, if measured, be better than what we observe in macaque IT cortex?  If the volume of cortical tissue is related to representational efficacy, it is likely that the human ventral stream would achieve even better performance.  While determining human homologues of macaque visual cortex is under active investigation, it is known that primary visual cortex in humans is twice as large as in macaque~\citep{VanEssen:2003vf}.  While this is suggestive that human visual representation may be even better under our metric, the scaling of human visual cortex over macaque may be optimizing representational aspects that we are not measuring here.  In summary, we suspect that the estimates for representational performance in macaque we have presented here provide a lower bound in performance of the human visual system.  One way to address human visual representation may be through the use of fMRI or inference of the human representational space from behavioral measurements.  We, and others in the neuroscience field, are actively pursuing these directions.

\subsection*{Where are we today?}
Under our analysis, we believe that the field has made significant advances with recent algorithms. On the intermediate level variation task (Medium Variation) these advances are quite evident: the recent representational learning algorithm in~\citep{le2011building} surpasses the representation in V4 and, surprisingly, the supervised algorithm of~\citep{Krizhevsky:2012wl} matches the representation in IT.  These advances are also evident on the high level variation task (High Variation): the \citep{le2011building} algorithm is narrowly better than V4 and the~\citep{Krizhevsky:2012wl} algorithm beats IT by an ample margin.  It will be informative to measure the elements of these models that lead to this performance and it will be interesting to see if purely unsupervised algorithms can achieve similar performance.

\subsection*{A vision for the future}
The methodology we have proposed here can be extended to other sensory domains where representation is critical and neural representations are thought to be effective.  For example, it should be possible to define similar task protocols for auditory stimuli and measure the neural responses in auditory cortex.  Such measurements would not only have implications for discovering effective auditory representations, but may also provide the data necessary to validate representational learning algorithms that are effective in multiple contexts.  Representational learning algorithms that prove effective across these domains may serve as hypotheses for a canonical cortical algorithm, a `holy grail' for artificial intelligence research.

\small{
\subsubsection*{Acknowledgments}
This work was supported by the U.S. National Eye Institute (NIH NEI: 5R01EY014970-09), the National Science Foundation (NSF: 0964269), and the Defense Advanced Research Projects Agency (DARPA: HR0011-10-C-0032).  C.F.C was supported by the U.S. National Eye Institute (NIH: F32 EY022845-01).  We thank Adam Coates, Quoc Le, and Alex Krizhevsky for their help in evaluating their models and comments on the paper.
}
\bibliographystyle{plainnat}
\small{
\bibliography{iclr2013}

\begin{thebibliography}{44}
\providecommand{\natexlab}[1]{#1}
\providecommand{\url}[1]{\texttt{#1}}
\expandafter\ifx\csname urlstyle\endcsname\relax
  \providecommand{\doi}[1]{doi: #1}\else
  \providecommand{\doi}{doi: \begingroup \urlstyle{rm}\Url}\fi

\bibitem[Baker et~al.(2002)Baker, Behrmann, and Olson]{Baker:2002bt}
CI~Baker, M~Behrmann, and CR~Olson.
\newblock {Impact of learning on representation of parts and wholes in monkey
  inferotemporal cortex}.
\newblock \emph{Nature Neuroscience}, 2002.

\bibitem[Braun(2006)]{Braun:2006va}
ML~Braun.
\newblock {Accurate Error Bounds for the Eigenvalues of the Kernel Matrix}.
\newblock \emph{JMLR}, 2006.

\bibitem[Braun et~al.(2008)Braun, Buhmann, and M{\"u}ller]{Braun:2008ul}
ML~Braun, JM~Buhmann, and KR~M{\"u}ller.
\newblock {On relevant dimensions in kernel feature spaces}.
\newblock \emph{JMLR}, 2008.

\bibitem[Canolty et~al.(2010)Canolty, Ganguly, Kennerley, Cadieu, Koepsell,
  Wallis, and Carmena]{Canolty:2010ts}
RT~Canolty, K~Ganguly, SW~Kennerley, CF~Cadieu, K~Koepsell, JD~Wallis, and
  JM~Carmena.
\newblock {Oscillatory phase coupling coordinates anatomically dispersed
  functional cell assemblies}.
\newblock \emph{PNAS}, 2010.

\bibitem[Churchland et~al.(2012)Churchland, Cunningham, Kaufman, Foster,
  Nuyujukian, Ryu, and Shenoy]{Churchland:2012bq}
MM~Churchland, JP~Cunningham, MT~Kaufman, JD~Foster, P~Nuyujukian, SI~Ryu, and
  KV~Shenoy.
\newblock {Neural population dynamics during reaching}.
\newblock \emph{Nature}, 2012.

\bibitem[Coates et~al.(2012)Coates, Karpathy, and Ng]{Coates:2012wm}
A~Coates, A~Karpathy, and A~Ng.
\newblock {Emergence of Object-Selective Features in Unsupervised Feature
  Learning}.
\newblock \emph{NIPS}, 2012.

\bibitem[DiCarlo et~al.(2012)DiCarlo, Zoccolan, and Rust]{DiCarlo:2012em}
JJ~DiCarlo, D~Zoccolan, and NC~Rust.
\newblock {How Does the Brain Solve Visual Object Recognition?}
\newblock \emph{Neuron}, 2012.

\bibitem[Fabre-Thorpe et~al.(1998)Fabre-Thorpe, Richard, and
  Thorpe]{FabreThorpe:1998te}
M~Fabre-Thorpe, G~Richard, and SJ~Thorpe.
\newblock {Rapid categorization of natural images by rhesus monkeys}.
\newblock \emph{Neuroreport}, 1998.

\bibitem[Felleman and Van~Essen(1991)]{felleman1991distributed}
DJ~Felleman and DC~Van~Essen.
\newblock {Distributed hierarchical processing in the primate cerebral cortex}.
\newblock \emph{Cerebral cortex}, 1991.

\bibitem[Fukushima(1980)]{Fukushima:1980iz}
K~Fukushima.
\newblock {Neocognitron: A self-organizing neural network model for a mechanism
  of pattern recognition unaffected by shift in position}.
\newblock \emph{Biological cybernetics}, 1980.

\bibitem[Hung et~al.(2005)Hung, Kreiman, Poggio, and DiCarlo]{hung2005fast}
CP~Hung, G~Kreiman, T~Poggio, and JJ~DiCarlo.
\newblock {Fast readout of object identity from macaque inferior temporal
  cortex}.
\newblock \emph{Science}, 2005.

\bibitem[Kiani et~al.(2007)Kiani, Esteky, Mirpour, and Tanaka]{Kiani:2007uz}
R~Kiani, H~Esteky, K~Mirpour, and K~Tanaka.
\newblock Object category structure in response patterns of neuronal population
  in monkey inferior temporal cortex.
\newblock \emph{Journal of Neurophysiology}, 2007.

\bibitem[Kobatake et~al.(1998)Kobatake, Wang, and Tanaka]{Kobatake:1998vc}
E~Kobatake, G~Wang, and K~Tanaka.
\newblock {Effects of shape-discrimination training on the selectivity of
  inferotemporal cells in adult monkeys}.
\newblock \emph{Journal of Neurophysiology}, 1998.

\bibitem[Kriegeskorte et~al.(2008{\natexlab{a}})Kriegeskorte, M, and
  P]{NikolausKriegeskorte:2008bz}
N~Kriegeskorte, Mur M, and Bandettini P.
\newblock {Representational Similarity Analysis -- Connecting the Branches of
  Systems Neuroscience}.
\newblock \emph{Frontiers in Systems Neuroscience}, 2008{\natexlab{a}}.

\bibitem[Kriegeskorte et~al.(2008{\natexlab{b}})Kriegeskorte, Mur, Ruff, Kiani,
  Bodurka, Esteky, Tanaka, and Bandettini]{Kriegeskorte:2008vz}
N~Kriegeskorte, M~Mur, DA~Ruff, R~Kiani, J~Bodurka, H~Esteky, K~Tanaka, and
  PA~Bandettini.
\newblock Matching categorical object representations in inferior temporal
  cortex of man and monkey.
\newblock \emph{Neuron}, 2008{\natexlab{b}}.

\bibitem[Krizhevsky et~al.(2012)Krizhevsky, Sutskever, and
  Hinton]{Krizhevsky:2012wl}
A~Krizhevsky, I~Sutskever, and G~Hinton.
\newblock {ImageNet classification with deep convolutional neural networks}.
\newblock \emph{NIPS}, 2012.

\bibitem[Le et~al.(2012)Le, Monga, Devin, Chen, Corrado, Dean, and
  Ng]{le2011building}
QV~Le, R~Monga, M~Devin, K~Chen, GS~Corrado, J~Dean, and AY~Ng.
\newblock {Building high-level features using large scale unsupervised
  learning}.
\newblock \emph{ICML}, 2012.

\bibitem[Li and DiCarlo(2010)]{Li:2010it}
N~Li and JJ~DiCarlo.
\newblock {Unsupervised Natural Visual Experience Rapidly Reshapes
  Size-Invariant Object Representation in Inferior Temporal Cortex}.
\newblock \emph{Neuron}, 2010.

\bibitem[Lowe(2000)]{lowe2000towards}
DG~Lowe.
\newblock {Towards a computational model for object recognition in IT cortex}.
\newblock In \emph{BMVC}, 2000.

\bibitem[Lowe(2004)]{lowe2004distinctive}
DG~Lowe.
\newblock {Distinctive Image Features from Scale-Invariant Keypoints}.
\newblock \emph{International Journal of Computer Vision}, 2004.

\bibitem[Majaj et~al.(2012)Majaj, Hong, Solomon, and DiCarlo]{Majaj:2012ui}
N~Majaj, H~Hong, E~Solomon, and JJ~DiCarlo.
\newblock {A unified neuronal population code fully explains human object
  recognition}.
\newblock In \emph{COSYNE}, 2012.

\bibitem[Montavon and M{\"u}ller(2012)]{Montavon:2012ub}
G~Montavon and KR~M{\"u}ller.
\newblock {Deep Boltzmann Machines and the Centering Trick}.
\newblock \emph{Neural Networks: Tricks of the Trade}, 2012.

\bibitem[Montavon et~al.(2011)Montavon, Braun, and M{\"u}ller]{Montavon:2011wp}
G~Montavon, ML~Braun, and KR~M{\"u}ller.
\newblock {Kernel Analysis of Deep Networks}.
\newblock \emph{JMLR}, 2011.

\bibitem[Mur et~al.(2012)Mur, Ruff, Bodurka, De~Weerd, Bandettini, and
  Kriegeskorte]{Mur:2012vq}
M~Mur, DA~Ruff, J~Bodurka, P~De~Weerd, PA~Bandettini, and N~Kriegeskorte.
\newblock {Categorical, Yet Graded -- Single-Image Activation Profiles of Human
  Category-Selective Cortical Regions}.
\newblock \emph{The Journal of Neuroscience}, 2012.

\bibitem[Oliva and Torralba(2007)]{Oliva:2007ui}
A~Oliva and A~Torralba.
\newblock {The role of context in object recognition}.
\newblock \emph{Trends in Cognitive Sciences}, 2007.

\bibitem[Orban(2008)]{orban2008higher}
GA~Orban.
\newblock {Higher order visual processing in macaque extrastriate cortex}.
\newblock \emph{Physiological Reviews}, 2008.

\bibitem[Pinto and Cox(2011)]{Cox:2011wj}
N~Pinto and D~Cox.
\newblock {Beyond simple features: A large-scale feature search approach to
  unconstrained face recognition}.
\newblock \emph{FG}, 2011.

\bibitem[Pinto et~al.(2008)Pinto, Cox, and DiCarlo]{Pinto:2008gj}
N~Pinto, David~D Cox, and JJ~DiCarlo.
\newblock {Why is Real-World Visual Object Recognition Hard?}
\newblock \emph{PLoS Computational Biology}, 2008.

\bibitem[Pinto et~al.(2009)Pinto, Doukhan, DiCarlo, and Cox]{Pinto:2009ho}
N~Pinto, D~Doukhan, JJ~DiCarlo, and DD~Cox.
\newblock {A High-Throughput Screening Approach to Discovering Good Forms of
  Biologically Inspired Visual Representation}.
\newblock \emph{PLoS Computational Biology}, 2009.

\bibitem[Pinto et~al.(2010)Pinto, Majaj, Barhomi, Solomon, and
  DiCarlo]{Pinto:2010hvm}
N~Pinto, N~Majaj, Y~Barhomi, E~Solomon, and JJ~DiCarlo.
\newblock {Human versus machine: comparing visual object recognition systems on
  a level playing field}.
\newblock In \emph{COSYNE}, 2010.

\bibitem[Pinto et~al.(2011)Pinto, Barhomi, Cox, and DiCarlo]{Pinto:2011iw}
N~Pinto, Y~Barhomi, DD~Cox, and JJ~DiCarlo.
\newblock {Comparing state-of-the-art visual features on invariant object
  recognition tasks}.
\newblock \emph{WACV}, 2011.

\bibitem[Riesenhuber and Poggio(1999)]{Riesenhuber1999}
M~Riesenhuber and T~Poggio.
\newblock {Hierarchical models of object recognition in cortex}.
\newblock \emph{Nature Neuroscience}, 1999.

\bibitem[Rosenblatt(1958)]{Rosenblatt:1958jc}
F~Rosenblatt.
\newblock {The perceptron: A probabilistic model for information storage and
  organization in the brain.}
\newblock \emph{Psychological review}, 1958.

\bibitem[Rust and DiCarlo(2010)]{Rust:2010uk}
NC~Rust and JJ~DiCarlo.
\newblock {Selectivity and tolerance (``invariance'') both increase as visual
  information propagates from cortical area V4 to IT}.
\newblock \emph{Journal of Neuroscience}, 2010.

\bibitem[Serre et~al.(2007)Serre, Wolf, Bileschi, Riesenhuber, and
  Poggio]{serre2007}
T~Serre, L~Wolf, S~Bileschi, M~Riesenhuber, and T~Poggio.
\newblock {Robust Object Recognition with Cortex-Like Mechanisms}.
\newblock \emph{PAMI}, 2007.

\bibitem[Sigala and Logothetis(2002)]{Sigala:2002ta}
N~Sigala and NK~Logothetis.
\newblock {Visual categorization shapes feature selectivity in the primate
  temporal cortex}.
\newblock \emph{Nature}, 2002.

\bibitem[Smola et~al.(1998)Smola, Sch{\"o}lkopf, and M{\"u}ller]{Smola:1998dq}
AJ~Smola, B~Sch{\"o}lkopf, and KR~M{\"u}ller.
\newblock {The connection between regularization operators and support vector
  kernels}.
\newblock \emph{Neural Networks}, 1998.

\bibitem[Stevenson and Kording(2011)]{Stevenson:2011ur}
IH~Stevenson and KP~Kording.
\newblock {How advances in neural recording affect data analysis}.
\newblock \emph{Nature Neuroscience}, 2011.

\bibitem[Stevenson et~al.(2012)Stevenson, London, Oby, and
  Sachs]{Stevenson:2012ux}
IH~Stevenson, BM~London, ER~Oby, and NA~Sachs.
\newblock {Functional Connectivity and Tuning Curves in Populations of
  Simultaneously Recorded Neurons}.
\newblock \emph{PLOS Computational Biology}, 2012.

\bibitem[Stringer and Rolls(2002)]{Stringer:2002hv}
SM~Stringer and ET~Rolls.
\newblock {Invariant Object Recognition in the Visual System with Novel Views
  of 3D Objects}.
\newblock \emph{Neural Computation}, 2002.

\bibitem[Van~Essen(2003)]{VanEssen:2003vf}
DC~Van~Essen.
\newblock {Organization of Visual Areas in Macaque and Human Cerebral Cortex}.
\newblock In \emph{The Visual Neurosciences}. 2003.

\bibitem[Wallis and Rolls(1997)]{wallis1997}
G~Wallis and ET~Rolls.
\newblock {Invariant Face and Object Recognition in the Visual System}.
\newblock \emph{Progress in Neurobiology}, 1997.

\bibitem[Wu et~al.(2006)Wu, David, and Gallant]{wu2006complete}
MC~Wu, SV~David, and JL~Gallant.
\newblock {Complete functional characterization of sensory neurons by system
  identification.}
\newblock \emph{Annual Review of Neuroscience}, 2006.

\bibitem[Zhang et~al.(2011)Zhang, Meyers, Bichot, Serre, Poggio, and
  Desimone]{Zhang:2011ut}
Y~Zhang, EM~Meyers, NP~Bichot, T~Serre, T~Poggio, and R~Desimone.
\newblock Object decoding with attention in inferior temporal cortex.
\newblock \emph{PNAS}, 2011.

\end{thebibliography}
}
\appendix{
\section*{Appendix}
\subsection*{Appendix A: High-throughput evaluation of the L3 model class}
%\subsection*{High-throughput evaluation of the L3 model class} 
%THIS WILL PROBABLY WANT TO GET TIGHTENED UP A BIT ... 
%While many hierarchical neural-network algorithms for object recognition share a common set of ``broad-stroke" properties, the performance of any one model class often depends strongly on the choice of parameters in a particular instantiation of that model -- e.g., the number of units per layer, the size of pooling kernels, exponents in normalization operations, etc.  While these details have often been presented as being incidental to the algorithm, recent work has shown that correctly setting these parameter choices is frequently critical to achieving robust real-world performance \citep{Pinto:2009ho}.  For this reason, it has useful to deploy high-throughput methods for optimization over model parameter spaces \citep{Pinto:2009ho, Begstra}.  In analogy to high-throughput screening approaches in molecular biology and genetics, these methods explore many thousands of potential network architectures and parameter instantiations, screening those that show promising robust object recognition performance for further development.  High-throughput methods have yielded significant, reproducible gains in performance across an array of basic object recognition tasks \citep{Cox:2011wj}.  
Using the L3 model class, we performed a high-throughput search of the parameter space by evaluating approximately 500 parameter selections at random on training image sets.  We ran each L3 model parameter instantiation on category-balanced subset of 12,800 randomly chosen images from the 128,000-image training set.  For each instantiation, we evaluated a kernel analysis protocol similar to that used for the testing set, but with the 16 object class labels of the training set as opposed to the 7 present in the testing set.  For each model instantiation, we also extracted features on the testing set images, and ran the standard kernel analysis protocol.

To evaluate the transfer between the training set and testing set, we examined how well training set scores predict testing set scores by comparing how relative performance rankings on the training set transfer to the testing set.  Figure~\ref{fig:L3_model} shows these results.  Performance on the training set is strongly correlated with performance on Medium ($r=0.64$) and High Variation ($r=0.58$) components of testing set, and weakly correlated on the Low Variation condition.  This might be expected, as the training set contains variation similar to the High Variation testing set.  The single best model from training achieves a high score on the testing set relative to other models in the training set and is in the range of the top machine-learning representations.  This data indicates that models that are trained using the provided training set can perform favorably on the testing set. 
%
%Given the distribution of training scores, we looked at two issues:  (i) the distribution of performances on the testing set produced by a randomly sampled population of three-layer hierarchical models, compared in an absolute sense in to other machine representations evaluated above, and (ii) how well training set scores predict testing set scores, comparing how relative performance ranking on the training set transfers to the testing set.   This allowed us to both determine the range of performances the L3 model class can achieve when trained on a task that is (at least in principle) similar to the one we evaluate in the classification protocol, and also obtain preliminary evidence that the training set we provide is actually informative for the testing set. 
%The distributions of performances over the L3 model space was highly varied, spanning a range from practically random to better than all but the best machine representations (fig. \ref{fig:L3_model}, top).  Performance on the training set is reasonably highly correlated with performance on Medium ($r=0.64$) and High Variation ($r=0.60$) components of testing set, though less well-correlated on the Low Variation component (fig \ref{fig:L3_model}, bottom).  The single best model from training achieves a high score on the testing set relative to other models in the training set and is the range of the top machine-learning representations.  This data indicates that models that are trained using the provided training set can perform favorably on the testing set. 
\begin{figure}[t]
\begin{center}
\includegraphics[width=.58\linewidth]{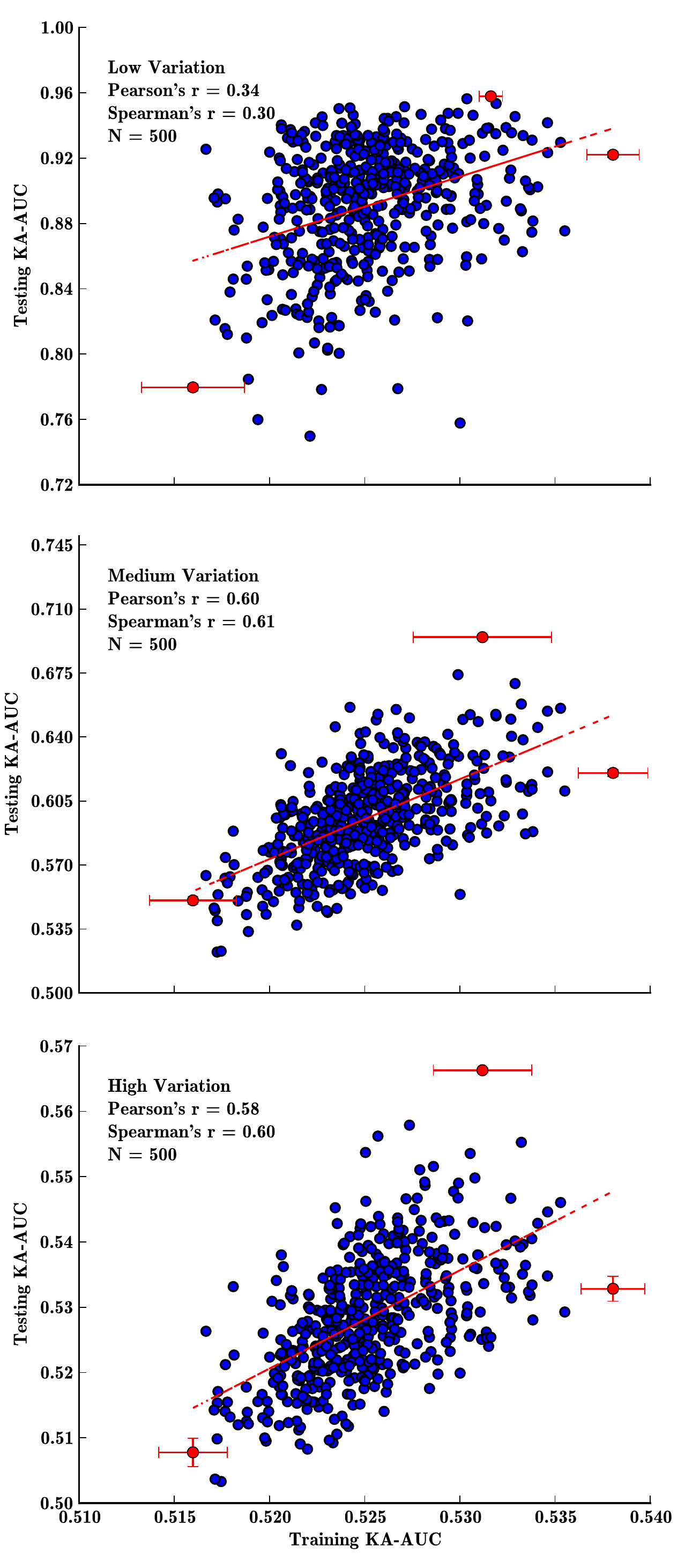}
\end{center}
\caption{High-throughput L3 model relationship between training and testing performance.  Each panel shows a scatter plot between the measured training KA-AUC and the testing KA-AUC for each variation level.  Red lines indicate best linear fit.  Red dots are the best and worst performing models on the training set and best performing model on each testing set (standard deviations are shown as error bars and are small in the testing axis).  Note that there is only one value for the training KA-AUC for each model.  The linear relationships we observe indicate that the provided training set is informative for the testing set.}
\label{fig:L3_model}
\end{figure}

\subsection*{Appendix B: Neural data collection}
We have collected neural data from V4 and IT across two adult male rhesus monkeys (\emph{Macaca mulatta}, 7 and 9 kg) by using a multi-electrode array recording system (BlackRock Microsystems, Cerebus System).  We chronically implanted three arrays per animal and have recorded the best 128 visually driven neural measurement sites (determined by separate pilot images) in one animal (58 IT, 70 V4) and 168 in another (110 IT, 58 V4).  During stimulus presentation we recorded multi-unit neural responses to our images (see Section \ref{sec:imageset}) from the V4 and IT sites.  Stimuli were presented on an LCD screen (Samsung, SyncMaster 2233RZ at 120Hz) one at a time.  Each image was presented for 100ms with a radius of 8$^\circ$ at the center of the screen on top of the half-gray background and was followed by a 100ms half-gray ``blank'' period.  The animal's eye movement was monitored by a video eye tracking system (SR Research, EyeLink II), and the animal was rewarded upon the successful completion of 6--8 image presentations while maintaining good eye fixation (jitter within $\pm2^\circ$ was determined acceptable fixation) at the center of the screen, indicated by a small (0.25$^\circ$) red dot.  Presentations with large eye movements were discarded.  In each experimental block, we recorded responses to all images of a certain variation level.  Within one block each image was repeated three times for Low Variation and once for Medium and High Variation.  This resulted in the collection of 28, 51, and 47 image repetitions for Low, Medium, and High Variation respectively.  All surgical and experimental procedures are in accordance with the National Institute of Health guidelines and the Massachusetts Institute of Technology Committee on Animal Care.

We convert the raw neural responses to a neural representation through the following normalization process.  For each image in a block, we compute the vector of raw firing rates across measurement sites by counting the number of spikes between 70ms and 170ms after the onset of the image for each site.  We then subtracted the background firing rate, which is the firing rate during presentation of a half-gray background or ``blank'' image, from the evoked response.  In order to minimize the effect of variable external noise, we normalize by the standard deviation of each site's response to a block of images for Medium and High Variation.  For the Low Variation stimuli, we divide the three repetitions within each block into three separate sets, each containing a complete set of images, and normalize by the standard deviation of each site's response within its set.  Finally, the neural representation is calculated by taking the mean across the repetitions for each image and for each site, producing a scalar valued matrix of neural sites by images.  This post-processing procedure is only our current best-guess at a neural code, which has been shown to account for human performance~\citep{Majaj:2012ui}.  Therefore, it may be possible to develop a more effective neural decoding, for example influenced by intrinsic cortical variability~\citep{Stevenson:2012ux}, or dynamics~\citep{Churchland:2012bq,Canolty:2010ts}.
%\begin{figure}[h]
%\begin{center}
%%\includegraphics[width=\linewidth]{figures/ICLR_Figure_1.pdf}
%\end{center}
%\caption{Schematic of experimental procedure and neural data preprocessing (this may not be necessary)}
%\end{figure}
%

\subsection*{Appendix C: KA-AUC and subsampling of neural sites}
Current experimental techniques only allow us to measure a small portion of the neurons in a cortical area.  We seek to estimate how our kernel analysis metric would be affected by having a larger neural sample.  In Figure~\ref{fig:neural_sampling} we estimate the effect of subsampling the neural population in our measurement, showing the KA-AUC as a function of the number of neural measurement sites.  To estimate the asymptotic convergence of each neural representation (V4 and IT) at each variation level, we fit a curve of the form $AUC(t) = a\! +\! b e^{-c t^d}$, where $t$ is the number of neural sites and $a$, $b$, $c$, and $d$ are parameters\footnote{We found that this functional form fit well a similar analysis performed on a computational representation in which we subsampled the number of features included in the analysis.  This allowed us to estimate the behavior of KA-AUC in much larger feature spaces ($>\! \!4000$ features) than in the neural measurements.}.  This provides us with an estimate of the KA-AUC for the entire neural population.  The estimated asymptotic values for the KA-AUC's for V4 are 0.89, 0.69, and 0.66, and for IT are 0.93, 0.91, and 0.75, for Low Variation, Medium Variation, and High Variation, respectively.  Interestingly we find that for the number of neural sites we have measured we are already approaching the asymptotic value.  Therefore, for the given task specification, preprocessing procedure, and convergence estimate, we believe we are reaching saturation in our estimate of KA-AUC for the neural population in V4 and IT.
% AUC = -A * np.exp(-.01* K * t**max((.05,X))) + 0.1*C
%V0 IT A: 1.29883173071 K: 100.735788757 X: 0.260014921987 C: 9.29527327456 @10k 0.929506630265 @100k 0.929527325041 @1M 0.929527327456
%V0 V4 A: 0.181644213406 K: 5.29786324095 X: 0.839385694893 C: 8.88503152225 @10k 0.888503152225 @100k 0.888503152225 @1M 0.888503152225
%V3 IT A: 268107602546.0 K: 2583.91068859 X: 0.024999983122 C: 9.05031300224 @10k 0.903031792042 @100k 0.904740347156 @1M 0.90499353245
%V3 V4 A: 0.24528552496 K: 15.9628375162 X: 0.547443051711 C: 6.86321445017 @10k 0.686321445012 @100k 0.686321445017 @1M 0.686321445017
%V6 IT A: 0.391553676161 K: 22.4803918134 X: 0.463745080964 C: 7.49689102093 @10k 0.749689062161 @100k 0.749689102093 @1M 0.749689102093
%V6 V4 A: 0.670856483309 K: 24.0525698013 X: 0.113247108551 C: 9.99999030036 @10k 0.661005523606 @100k 0.72338307833 @1M 0.787550992136
\begin{figure}[t]
\begin{center}
\includegraphics[width=.6\linewidth]{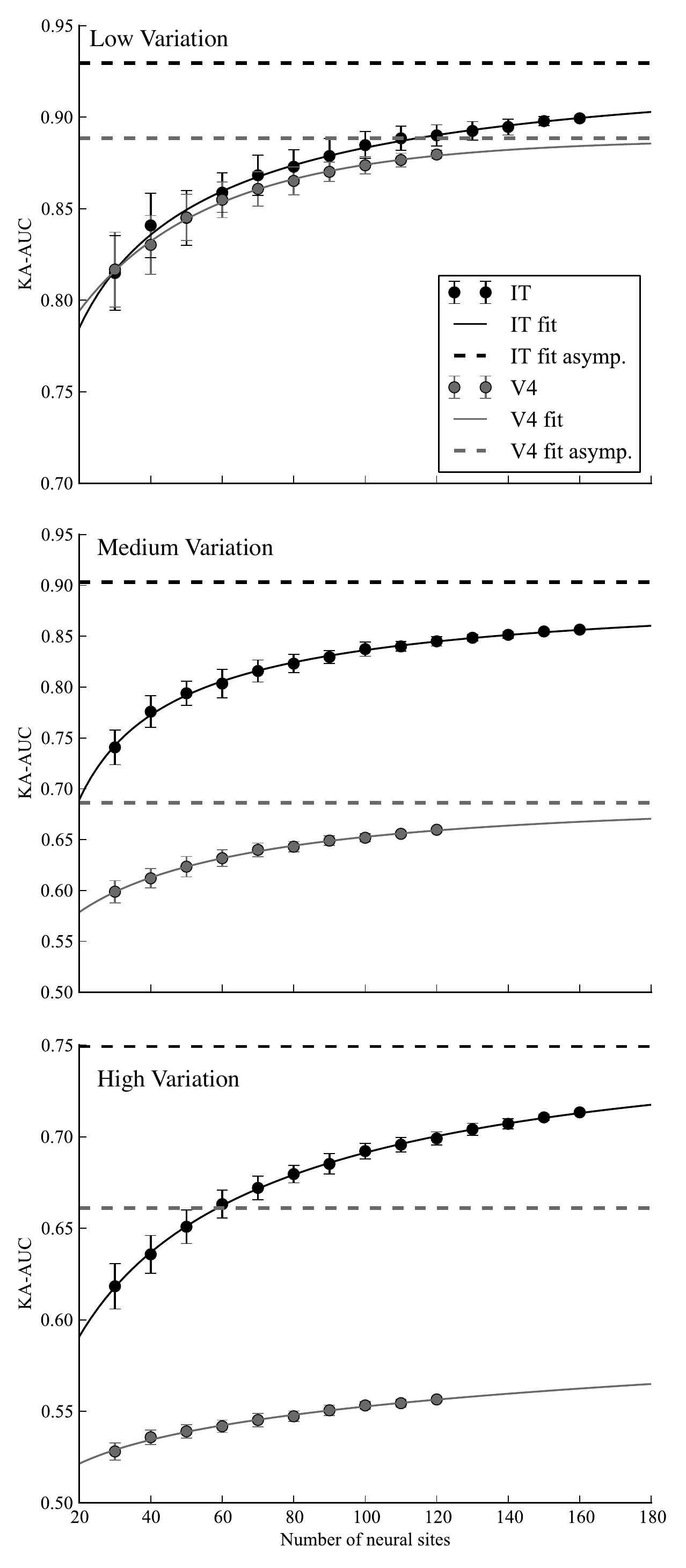}
\end{center}
\caption{Effect of sampling in neural areas.  We estimate the effect of sampling the neural sites on the testing set KA-AUC.  Each panel shows the effect for each variation level.  Best fit curves are shown as solid lines with measured samples indicated by filled circles.  Estimated asymptotes are indicated by dashed horizontal lines.  See text for more details.}
\label{fig:neural_sampling}
\end{figure}
}%end \appendix

\small{
\subsubsection*{Errata}
An earlier version of this manuscript contained incorrectly computed kernel analysis curves and KA-AUC values for V4, IT, and the HT-L3 models.  They have been corrected in this version.
}
\end{document}